\documentclass[submission,copyright,creativecommons]{eptcs}
\providecommand{\event}{SOS 2007} 
\usepackage{breakurl}             
\usepackage{underscore}           

\usepackage{graphicx}
\usepackage{cite}

\usepackage{xspace}
\usepackage{listings}

\usepackage{hyperref}
\usepackage{paralist}
\usepackage{footmisc}
\usepackage{pdfsync}
\usepackage{srcltx}

\usepackage[inline]{enumitem}

\usepackage{listings}
\lstdefinestyle{asp-style}{
	language=Prolog,
	frame=lines,
	keywordstyle=\linespread{1.1}\small\ttfamily,
	basicstyle=\linespread{1.1}\small\ttfamily,
	breaklines=true,
}
\lstdefinestyle{csharp-style}{
	language=[Sharp]C,
	frame=lines,
	keywordstyle=\linespread{1.1}\small\ttfamily,
	basicstyle=\linespread{1.1}\small\ttfamily,
	breaklines=true,
}
\usepackage[colorinlistoftodos,textwidth=80pt]{todonotes}
\presetkeys{todonotes}{color=blue!20, bordercolor=blue!80}{}

\newcommand{\embasp}{\protect\textsc{EmbASP}\xspace}

\newcommand{\unity}{Unity\xspace}

\newcommand{\quo}[1]{``#1''}

{%
	\begin{displaymath}%
	\begin{array}{l}%
}%
{%
	\end{array}%
	\end{displaymath}%
}%

\usepackage[acronym,nonumberlist,nohypertypes={acronym}]{glossaries}

\newacronym{ai}{AI}{\textit{Artificial Intelligence}}
\newacronym{asp}{\textsc{ASP}}{\textit{Answer Set Programming}}
\newacronym{pddl}{\textsc{PDDL}}{\textsc{Planning Domain Definition Language}}
\newacronym{cgi}{CGI}{Computer-Generated Imagery}

\title{Reasoning in Highly Reactive Environments}
\author{Francesco Pacenza
\institute{Department of Mathematics and Computer Science\\
University of Calabria\\
Rende, Italy}
\email{pacenza@mat.unical.it}
\and
Advisor: Giovambattista Ianni
\institute{Department of Mathematics and Computer Science\\
University of Calabria\\
Rende, Italy}
\email{\quad ianni@mat.unical.it}
}

\def\titlerunning{Reasoning in Highly Reactive Environments}
\def\authorrunning{F. Pacenza}
\begin{document}
\maketitle


\section{Introduction and problem description}
The aim of my Ph.D. thesis concerns {\em Reasoning in Highly Reactive Environments}.

As {\em reasoning in highly reactive environments}, we identify the setting in which a knowledge-based agent, with given goals, is deployed in an environment subject to repeated, sudden and possibly unknown changes. This is for instance the typical setting in which, e.g., artificial agents for video-games (the so called \quo{bots}), cleaning robots, bomb clearing robots, and so on are deployed. In all these settings one can follow the classical approach in which the operations of the agent are distinguished in \quo{sensing} the environment with proper interface devices, \quo{thinking}, and then behaving accordingly using proper actuators~\cite{DBLP:conf/ruleml/2017s}.

In order to operate in an highly reactive environment, an artificial agent needs to be:
\begin{itemize}
	\item \textit{Responsive} $\rightarrow$ The agent must be able to react repeatedly and in a reasonable amount of time;
	\item \textit{Elastic} $\rightarrow$ The agent must stay reactive also under varying workload;
	\item \textit{Resilient} $\rightarrow$ The agent must stay responsive also in case of internal failure or failure of one of the programmed actions in the environment.
\end{itemize}
Nowadays, thanks to new technologies in the field of Artificial Intelligence, it is already technically possible to create AI agents that are able to operate in reactive environments. Nevertheless, several issues stay unsolved, and are subject of ongoing research.

\section{Goal of the research}
With the term {\em data streams} we identify an unbounded sequences of time-varying data elements, that means a \quo{continuous} flow of information. The assumption is that recent information are more relevant as they describe the current state of a dynamic system.
In the latest years, due to the increasing volume of data streams and the crucial time requirements of many applications, different real-time approaches have been studied in order to ensure that a query answer is pushed as soon as it becomes available to the consumers. Notably, in the field of stream reasoning, some work aim at introducing advanced reasoning capabilities under a well understood and formalized framework~\cite{EiterLars, EiterTicker}. By the way, {\em Stream Reasoning} is a quite young and unexplored area, and therefore many different issues need to be solved.
To this end we aim at extending the functionality of the already existing tools in order to give them the possibility to work also in highly reactive environment in a context in which the amount of data coming each second is relatively small, yet time requirements are very strict. In order to achieve our goal we focus on the usage of the pure \acrfull{asp} semantics in a repeated evaluation setting. In this respect, we aim to study incremental techniques able to reuse previously computed inferences, allowing to knock down the time required during the reasoning phase. We aim to achieve this goal particularly focusing on the optimization of the grounding step.
 In addition to this, we would like to develop also an incremental answer set solver able to evaluate solutions at runtime in subsequent, similar, evaluation shots. Moreover, we are currently working on a framework able to ease the development of AI-agents in highly reactive contexts such as video-games and real robots. Some of these results have been already achieved and will be briefly described in sections~\ref{results}.


\section{Background and overview of the existing literature}
\label{background}
During the last years, some different approaches have been explored in order to create responsive agents that are able to operate in highly reactive environments. Most of them make use of \textit{deep learning techniques} in order to let the agent be able to emulate human learning. These techniques are applied in the field of video-games for automatic game playing accordingly to~\cite{DeepLearning}. The same technique can be also used for predicting human behavior as discussed in~\cite{DeepLearningHuman}. This kind of approaches have their \quo{limits} especially when one aims to {\em a)} generalize reasoning (indeed, deep learning techniques are usually built ad-hoc for a specific operating environment); {\em b)} create development and reasoning tools that can be \quo{used} as middleware for the production of the so called \quo{bots}, in which \quo{intelligent} capabilities can be tuned, refined and prototyped; and {\em c)} provide intelligible explanations of choices taken by the agent.
\vspace{2mm}
\\In the \textit{Answer Set Programming (ASP)} community there have been some recent studies about reasoning in reactive environments; in particular, researchers recently proposed ASP-based forms of reactive reasoning. One of these (called \quo{\textit{oclingo}}) allows to implement real-time dynamic systems running \underline{online} in changing environments (as described in~\cite{ReactiveASP123}).
This new technology paves the way for applying ASP in many areas like robots, bots, and video-games due to the fact that now an ASP solver can be also adopted for online usage. However \textit{oclingo} cannot be defined as totally \quo{declarative}. Indeed, the user has to specify how the environment will evolve and it is necessary to specify the operational design of some \textit{base}, \textit{cumulative} and \textit{volatile} groups of rules, that should be respectively evaluated only one time, at each iteration and only in one iteration before being discarded.
\vspace{2mm}
\\A second promising ASP extension like {\em HEX}~\cite{HexProgram} provides the possibility of a bidirectional access to external sources of knowledge and/or computation using the concept of external atoms; the extension {\em ActHEX}~\cite{ActHEX} of HEX programs introduces the notion of action atoms, which are associated to corresponding functions capable of actually changing the state of external environments. Using this ASP extension it is in practice possible to realize artificial agents able to operate in many different settings (for instance, an HEX extension has been successfully used for implementing an AI agent operating in video-games scenarios as described in~\cite{AngryHEX}). However, ActHex does not
make explicitly possible to check and act if the planned actions in the schedule have failed at each shot. Moreover, in very complex scenarios, the implementation of ActHex knowledge bases could become very hard and unhandy due to the complex ways for programming the order of actions composing a scheduled plan.
Furthermore, we found in literature a large variety of languages designed for programming logical Agents and Multi-Agent Systems in contexts very similar to our setting. One of these is Jason~\cite{jason}; Jason is a platform for developing agents based on the Agent-Speak logic language. A Jason agent consists of a set of plans, each of which has a triggering event, a context, and a body of actions. Moreover also other languages like IMPACT~\cite{IMPACT}, DALI~\cite{DALI}, ALP~\cite{alp} and ASTRA~\cite{Astra} can be used to design logical Agents and Multi-Agent Systems.

\section{Current status of the research}
In the latest years, also the \acrlong{asp} community focused on {\em Stream Reasoning}. Implementing stream reasoning with a typical \acrshort{asp} system requires to start each computation from scratch again and again at each evaluation shot. Indeed, the typical structure of an ASP system mimics such way of defining the semantics by relying on a grounding module that takes in input a non-ground logic program $P$ and produces an equivalent propositional theory $gr(P)$; the grounder is coupled with a subsequent solver module that applies proper resolution techniques on $gr(P)$ for computing the actual semantics in form of {\em answer sets} $AS(gr(P))$~\cite{DBLP:journals/aim/KaufmannLPS16}.

In the context of stream reasoning and multi-shot reasoning engines~\cite{EiterTicker, DBLP:journals/ai/BrewkaEGKLP18, DBLP:journals/tplp/GebserKKS19} based on answer set programming, a quite typical setting is when the grounding step is repeatedly executed on slightly different input data, while a short computation time window is allowed. In this respect, we decided to focus our attention on the following open issues:
\begin{enumerate}
  \item reuse as much as possible of the previously computed knowledge in logic programs;
  \item knocking down the time required for the evaluation of logic programs in subsequent and similar evaluation shots;
  \item stop and restart the computation of the answer sets when new facts are fed in input.
\end{enumerate}
Concerning points 1 and 2 we focused on the usage of the pure ASP semantics in a repeated evaluation setting.
In particular, we aimed at closing some of the mentioned gaps in the current state-of-the-art by investigating the possibility of generating incrementally larger ground programs for a fixed non-ground logic program. In our recent experiments, we showed that this approach, which we called \quo{overgrounding}, pays off in terms of performance.
Overgrounded programs can be reused in combination with deliberately many different sets of inputs; this gives less control on the computational procedure easing the burden of taking care of it: indeed, the overgrounding process and the multi-shot machinery can be completely transparent to the user.
We expect this setting to be particularly favourable when non-ground input knowledge bases are constituted of small set of facts, typical of declaratively programmed video game agents, or robots. Part of this work has been published in~\cite{CILC:2019Pacenza} while its extension will be published in~\cite{ICLP:2019Pacenza}.

\section{Preliminary results}
\label{results}
\subsection{First year}
In order to achieve our goal, we started to investigate the literature~\cite{HexProgram, EiterLars, EiterTicker, AngryHEX, ActHEX, DeepLearningHuman, DeepLearning, Replanning2, TimeEstimation, ReactiveASP123}. Then, we focused our research on artificial agents for video-games (also called \quo{bots}). We worked on embedding rule-based {\em Reasoning Modules} into the well-known Unity\footnote{\url{https://unity3d.com/unity}} game development engine. To this end, we presented an extension of \embasp\footnote{\url{https://www.mat.unical.it/calimeri/projects/embasp}}, a framework to ease the integration of declarative formalisms with generic applications. Finally, we proved the viability of our approach\footnote{All the development material, including logic programs, source code and a fully playable version of the game are available at \\ \url{https://github.com/DeMaCS-UNICAL/Pac-Man-Unity-EmbASP}.} by developing a proof-of-concept Unity game that makes use of ASP-based AI modules~\cite{aspVideoGame}.

As a side research topic we investigated to what extent the ASP-based approach is scalable enough for industrial contexts in the field of video games, by proposing a Unity extension capable to automatically generate dungeons maps\footnote{Both versions of our prototypes, together with logic program specifications and source code are fully available online at \\ \url{https://github.com/DeMaCS-UNICAL/DCS-Maze_Generator-GVGAI} and \\ \url{https://github.com/DeMaCS-UNICAL/DCS-Maze_Generator-Unity}}. In this context we first investigated over the usage of a partition-based generation technique~\cite{generationGames}, then we proposed a multiple step-generation approach, set in the context of the 2-D caves generation domain, where each step is declaratively controlled by an ASP specification. With respect to existing literature~\cite{Smith, Smith2}, our approach promises to be better scalable to real contexts with higher size mazes; experiments aimed at confirming that are currently ongoing. Finally we developed two plugins based on our generation technique, which were respectively deployed as an asset available in the Unity development and in the GVGAI~\cite{gvgai} frameworks~\cite{aspVideoGame}.

\subsection{Second year}
During the currently ongoing second year of studies, we continued to work on the lines of research described above. In particular we extended the integrated rule-based {\em Reasoning Module} framework by adding some new functionalities in order to ease the integration of declarative formalisms within the typical game development workflow in \unity. We solved some of the issues encountered in the previous work as, for example, the possibility to execute asynchronous tasks in a \unity game and the possibility to share the internal \unity data structure with the AI module. Thanks to these improvements we give the possibility to developers to directly attach a reasoning task to a trigger condition on an object property or at scheduled times.

After some preliminary studies we focused our attention on the usage of the pure ASP semantics in a repeated evaluation setting. We characterized a class of ground programs called {\em embeddings} or {\em embedding programs} which introduce a model-theoretic-like notion of ground programs; second, we proposed an incremental grounding strategy able to reuse previously grounded programs. This technique allows to knock down the time necessary for performing the instantiation of logic programs in subsequent, similar, evaluation shots. In particular, we maintain a stored ground logic program which grows monotonically from one shot to another; such {\em overgrounded programs} are series of embedding programs that become more and more general while elaborating consecutive shots, increasingly adding potentially useful rules. Cached rules can be reused in combination with deliberately many different sets of inputs. Then we developed a first implementation of an intelligent grounder able to apply the new introduced differential algorithm and finally tested our architecture over some applicative domains. We obtained very promising results.

Currently we are performing other tests using the benchmark suite used during the ASP Competition.
Part of this work has been published in~\cite{CILC:2019Pacenza} while its extension will likely be published in~\cite{ICLP:2019Pacenza}.

\section{Open issues and expected achievements}
In the future work we aim at solving some of the issues encountered during the last 2 years. Some of them are reported in the following:
\begin{enumerate}
    \item consider simplified ground programs;
    \item introduction of memory constraints during the grounding step;
    \item development of an incremental solver considering also the model generation phase;
    \item possibility to stop and restart the computation of an answer set.
\end{enumerate}
Studying the first issue is of great importance because including simplifications in ground programs would knock down the usage of computational resources in terms of memory consumption; indeed, storing a non simplified ground program is more expensive than storing a simplified yet general ground program due to its larger and increasing size in terms of number of stored rules. Our next step is to add the simplification step in our incremental grounder framework.

After the implementation of the simplification, another issue we will deal with is the possibility to add a memory constraint during the grounding stage. Thanks to this possible improvement, one could set an upper bound to memory consumption. One possibility could be to remove the less triggered rules (or simply the oldest ones) when the memory limit has been exceeded.

Concerning the third issue, this is one of the crucial points that we need to solve in order to release a monolithic system able to directly produce incremental answer sets. At this stage of research activities, we are performing benchmarks also on the solving time of the overgrounded programs that are obviously greater than the solving time of the simplified instances. This time could be knocked down if a solver could be fed at each shot with only the newly computed ground rules. In this case it is not necessary to read at each iteration the entire ground program but only the new incremental rules.

Finally, we aim at adding the possibility to stop and restart the computation of the answer sets when the system understands that the current knowledge base has been changed by adding or removing assertions, so the set of candidate solutions is no more the same and it is needed to restart the computation over the new set of facts.

\section{Conclusions}
In this work we are investigating the field of stream reasoning with specific focus on achieving high performance. Our goal is to extend the currently developed framework to ease the generation of responsive, elastic and resilient artificial agents. Here we reported on how we are proceeding in order to make this goal possible and so we have presented our current state of the work. In the last year we worked on the implementation of an incremental grounder able to apply a differential algorithm in subsequent and similar shots over different input data; currently we are working on an improvement of the incremental grounder in order to take in consideration simplifications rules that can allow to produce smaller programs.
Moreover, we have presented some of the research themes that we would like to deepen (add a memory consumption limit, develop of a full incremental solver, introduce the possibility to restart a computation at runtime).

\bibliographystyle{eptcs}

\end{document}